\documentclass[conference]{IEEEtran}
\IEEEoverridecommandlockouts

\usepackage{gensymb}
\usepackage{url}
\usepackage{algorithm}
\usepackage{hyperref}

\usepackage{subcaption}
\usepackage{array}
\usepackage{pifont}

\usepackage{enumitem}
\usepackage{xspace}
\usepackage{verbatim}
\usepackage{wrapfig}
\usepackage{lipsum}
\usepackage{arydshln}

\usepackage{cite}
\usepackage{amsmath,amssymb,amsfonts}
\usepackage{algorithmic}
\usepackage{graphicx}
\usepackage{textcomp}
\usepackage{multirow}
\usepackage{xcolor}
\usepackage{tabularx}

\def\BibTeX{{\rm B\kern-.05em{\sc i\kern-.025em b}\kern-.08em
    T\kern-.1667em\lower.7ex\hbox{E}\kern-.125emX}}
    
\begin{document}

\title{TileTracker: Tracking Based Vector HD Mapping using Top-Down Road Images\\
\thanks{This project was conducted during an internship at NVIDIA Corporation}
}

\author{\IEEEauthorblockN{Mohammad Mahdavian}
\IEEEauthorblockA{\textit{Department of Computing Sciences} \\
\textit{Simon Fraser University}\\
Burnaby, Canada \\
mmahdavi@sfu.ca}
\and
\IEEEauthorblockN{Mo Chen}
\IEEEauthorblockA{\textit{Department of Computing Sciences} \\
\textit{Simon Fraser University}\\
Burnaby, Canada \\
mochen@cs.sfu.ca}
\and
\IEEEauthorblockN{Yu Zhang}
\IEEEauthorblockA{\textit{NVIDIA Corporation} \\
Santa Clara, United States of America \\
yuzh@nvidia.com}
}

\maketitle

\begin{abstract}

In this paper, we propose a tracking-based HD mapping algorithm for top-down road images, referred to as tile images. While HD maps traditionally rely on perspective camera images, our approach shows that tile images can also be effectively utilized, offering valuable contributions to this research area as it can be start of a new path in HD mapping algorithms. We modified the BEVFormer layers to generate BEV masks from tile images, which are then used by the model to generate divider and boundary lines. Our model was tested with both color and intensity images, and we present quantitative and qualitative results to demonstrate its performance.

\end{abstract}

\begin{IEEEkeywords}

HD mapping, top-down images, tracking
\end{IEEEkeywords}

\section{Introduction}

Autonomous vehicles rely on accurate environmental data from various sensors, including cameras, lidar, radar, IMUs, and GPS. These sensors work together, with their data fused to generate precise commands for vehicle navigation~\cite{yeong2021sensor}.

Some methods estimate waypoints and navigational commands directly from sensor data in an end-to-end manner~\cite{agand2023letfuser, Chitta2022PAMI, shao2023safety}. Large Visual Models (LVMs) derive actions from camera images~\cite{chen2024driving}, while Reinforcement Learning (RL) optimizes navigation based on reward signals~\cite{gutierrez2022reinforcement}.

High Definition (HD) maps serve as a valuable and detailed resource for navigating autonomous vehicles. To create these maps for various areas and cities, vehicles equipped with multiple sensors, such as cameras, lidar, and IMUs, traverse the roads, capturing necessary data. Human annotators then identify fixed road features like lane dividers, traffic lights, pedestrian crossings, and road boundaries. Following this, deep learning models have been employed to learn and estimate these road features for inclusion in the HD map based on the sensor data~\cite{liao2022maptr, liao2023maptrv2,yuan2024streammapnet, chen2024maptracker}.
One well-known method for HD mapping is MapTR~\cite{liao2022maptr, liao2023maptrv2}, which offers an end-to-end, transformer-based~\cite{vaswani2017attention}, unified permutation-equivalent modeling approach. This method models each map element as a point set with a group of equivalent permutations, enabling accurate shape descriptions and stabilizing the learning process. However, MapTR generates road features on a frame-by-frame basis, which can result in noisy and inconsistent HD maps.
To address these issues, StreamMapNet~\cite{yuan2024streammapnet} utilizes long-sequence temporal modeling of video data. It employs multi-point attention and temporal information to produce large-scale local HD maps with enhanced stability. Additionally, MapTracker~\cite{chen2024maptracker} approaches the mapping problem as a tracking task, leveraging a memory of latents to ensure consistent reconstructions over time. This method adds a sensor stream into memory buffers of two latent representations: raster latents in bird's-eye-view (BEV) space and vector latents over road elements.

In these types of HD mapping methods, local HD maps are typically generated from perspective view (PV) images. This approach can limit model performance due to the need for PV-to-BEV conversion. In this work, we address this issue by developing HD maps based on MapTracker using top-down road images, known as tile images. We utilize NVIDIA's road dataset to generate sequences of these top-down tile images. Additionally, we modify MapTracker to create BEV segmentation directly from the tile images, enabling the algorithm to generate vectors for local HD maps. These local maps are then integrated to produce comprehensive global HD maps.
\vspace{-1mm}

\section{Methodology}

In this section, we explain our algorithm. As mentioned before, our approach is derived from MapTracker~\cite{chen2024maptracker} and we have made modifications on this method to be able to generate local HD maps from top-down tile images.

This method employs a robust memory mechanism to accumulate sensor data into two distinct memory representations. The first is a top-down BEV memory of the vehicle's surrounding area, and the second is a vector memory for road elements. At each frame, the algorithm selects a subset of past latent memories for fusion. The vector memory mechanism also uses tracking methods to maintain a sequence of memory latents associated with each road element.
In the BEV module, the BEV memory buffer $M_{BEV}(t-1)$ generates the initial BEV features for each frame. Then, a deformable self-attention mechanism enhances the BEV memory features. In the next step, for MapTracker, a spatial deformable cross-attention layer, adapted from BEVFormer~\cite{li2022bevformer}, converts perspective view image features $I(t)$ into BEV features $M_{BEV}(t)$. 

We have modified the BEVFormer layer to generate BEV features from top-down tile images by utilizing top-down camera intrinsic and extrinsic matrices. In general, the BEVFormer processes six camera image features along with their respective camera matrices and related information. Deformable attention~\cite{zhu2020deformable} is used to integrate the image features into the corresponding areas of the BEV mask.
In our approach, the extrinsic camera matrix is constructed by combining the vehicle's translation and rotation matrices, followed by an additional downward rotation to obtain the final extrinsic matrix. In this way, the deformable attention mechanism integrates the top-down tile image features with the entire BEV mask, leading to efficient generation of the BEV mask.

In the next layer, latent features from previous frames that are stored in a buffer get fused with the current frame's features to create the final BEV mask, $M_{BEV}(t)$. The algorithm uses a strided selection of four memories, corresponding to various vehicle positions relative to the current position, for the fusion. 

For vector predictions, a transformer-based tracking approach is used to propagate vector queries, combined with new learnable element candidates to generate vector features for new frames. Subsequently, a standard self-attention mechanism, along with BEV-to-Vector cross-attention and vector memory fusion, is employed to estimate vectors, $Y(t)$, containing vector points $p$ for the next frame.

\begin{figure}
\begin{subfigure}{.5\textwidth}
    \centering
    \includegraphics[width=0.95\linewidth,trim={0cm  0cm 0cm 0.0cm },clip]{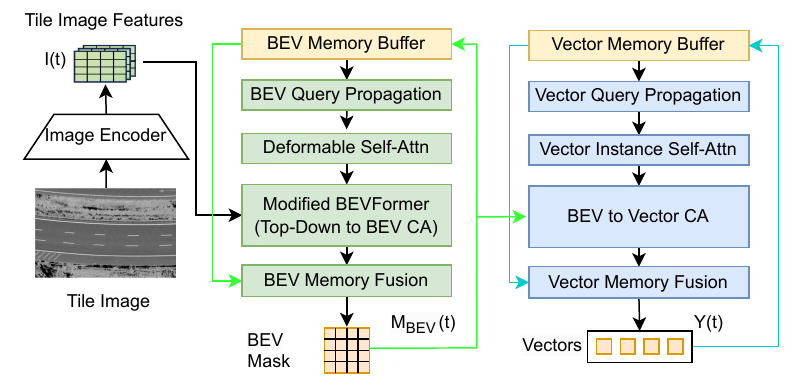}  
\end{subfigure}\hspace{0mm} \vspace{-5mm}

\caption{Main TileTracker model structure. Green and blue boxes show BEV and vector modules, respectively.}
\label{fig:main}\vspace{-5mm}
\end{figure}

\section{Training}

The road elements defined in our dataset include divider and boundary lines. We extract top-down tile images with the size of ($H,W$) that includes intensity and color images as well as the ground truth (GT) for each line found in the image. We train two separate model with using intensity and color images as our model inputs to investigate their performance. Additionally, we divide all the line types (solid, double solid, dashed, etc.) into two main categories, divider and boundary lines. Each road element in the GT is denoted as $\hat{Y} = (\hat{V},\hat{c})$ where $\hat{V}$ is 20 points interpolated from the raw vector and $\hat{c}$ specifies the line type. Also, the lines are rasterized into a segmentation image using OpenCV and PIL libraries to generate GT for the BEV mask, $\hat{S}$.

\subsection{Loss Functions}

The loss functions used for training our model, focus on the BEV mask and vector generation. For the BEV masks, we use per-pixel Focal loss~\cite{lin2017focal} and per-class Dice loss~\cite{milletari2016v} between the predictions and the GT defined by 

\vspace{-4mm}

  \begin{align}
    \mathcal{L}_\text{BEV} = \lambda_1 \mathcal{L}_\text{focal}(S(t),\hat{S}(t)) + \lambda_2 \mathcal{L}_\text{dice}(S(t),\hat{S}(t))
 \end{align}

For the vector generation, we use a tracking style loss which consists of a focal loss, $\mathcal{L}_\text{focal}$, and a permutation-invariant line coordinate loss, $\mathcal{L}_\text{line}$. Their combination defines $\mathcal{L}_\text{track}$ as

\vspace{-4mm}

  \begin{align}
    \mathcal{L}_\text{track} = \lambda_3 \mathcal{L}_\text{focal}(p, \hat{c}) + \lambda_4 \mathcal{L}_\text{line}(V(t),\hat{V}(t))
 \end{align}

We also borrow the transformation loss, $\mathcal{L}_\text{trans}$, from StreamMapNet to enforce the query transformation to maintain the vector shape and line type. Therefore the complete loss function would become

\vspace{-4mm}

  \begin{align}
    \mathcal{L} = \mathcal{L}_\text{BEV} + \mathcal{L}_\text{track} + \lambda_5 \mathcal{L}_\text{trans}
 \end{align}

\subsection{Model Training Details}

The training has three main stages. 1) Pre-training the image backbone and BEV encoder to generate BEV masks, 2) Warming up the vector decoders while keeping the remaining model components frozen, 3) Training the complete model structure.

The loss weights are set as follow: $\lambda_1 = 10.0$, $\lambda_2 = 1.0$, $\lambda_2 = 5.0$, $\lambda_2 = 50.0$ and $\lambda_5 = 0.1$. We use AdamW as our optimizer and initial learning rate of 5e-4 which is decreased to 1.5e-6 using a cosine learning rate scheduler.

\section{Experiments}

We train our model using 8 NVIDIA V100 GPUs and the three stages need 12, 4 and 24 epochs, respectively. Also the batch size of 4 is considered for all training stages. In this section, we explain our dataset generation and preprocessing.

\subsection{Dataset} 

\begin{figure}

\begin{subfigure}{.24\textwidth}
    \centering
    \includegraphics[width=0.95\linewidth,trim={0cm  0cm 0cm 0cm },clip]{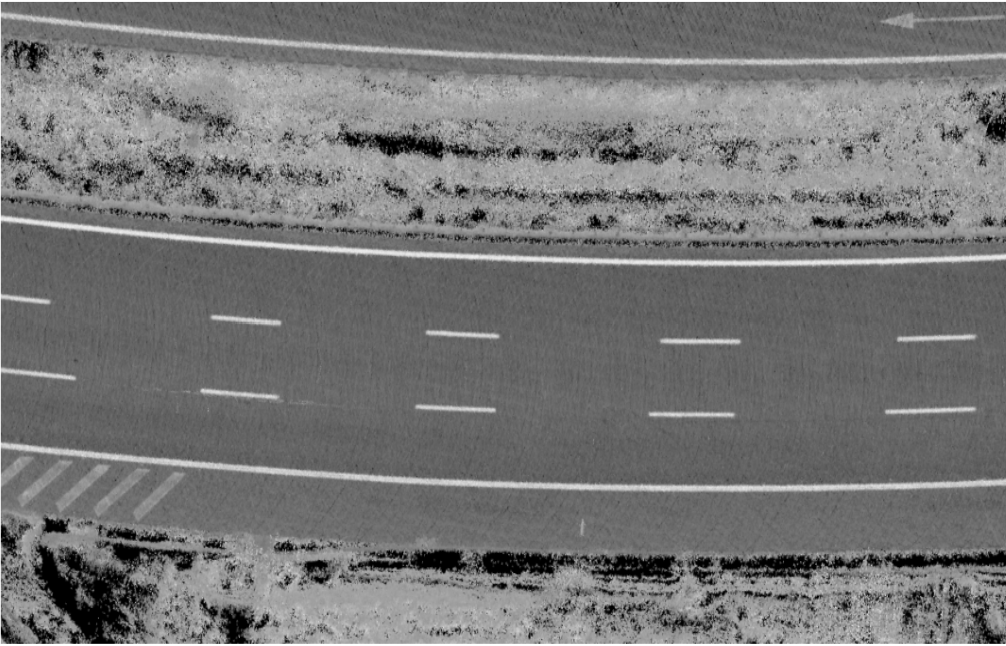}  
\end{subfigure}\vspace{0mm}
\begin{subfigure}{.24\textwidth}
    \centering
    \includegraphics[width=0.95\linewidth,trim={0cm  0cm 0cm 0.0cm },clip]{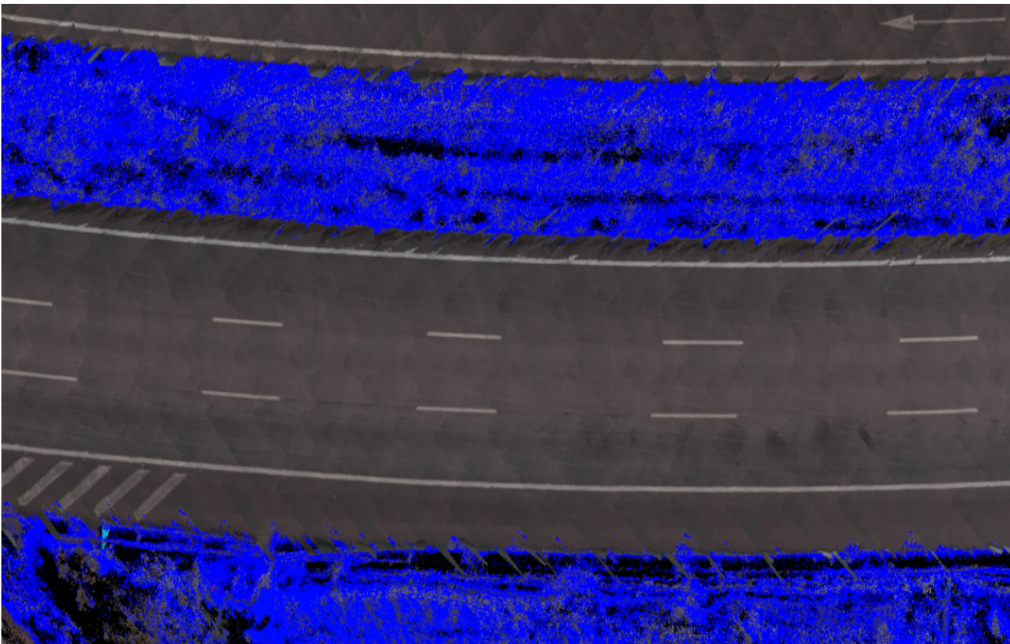}  
\end{subfigure}\hspace{0mm} 

\caption{An example of intensity (left) and color (right) images is shown. These images are generated from 3D colored lidar point clouds and displayed as 2D top-down BEV images, with intensity represented in grayscale and color images in their original colors.}

\label{fig:images}\vspace{-5mm}
\end{figure}

We use data collected from various roads and tracks to train our model. This dataset includes top-down tile intensity and color images covering a 30$\times$60 meters area, GT for different divider line types and boundary lines within the images, and the car's GPS location. 
In the data generation process, a 3D colored LiDAR system was employed to capture track data. The resulting color and intensity images provide 2D top-down views, displaying colored and grayscale representations of the captured points. A sample of them is provided in Fig.~\ref{fig:images}.

It is crucial to preprocess the dataset to ensure it is suitable for model training. The input data must be continuous with reasonable frame differences, as our tracking algorithm relies on finding similarities between consecutive frames. Consequently, the tracks are generated so that two consecutive frames have a minimum distance of 1 meter, though this is not always guaranteed. In some instances, the distance between frames may exceed 5 meters, in which case we split them into separate tracks. We only consider tracks with at least 20 frames and limit each track to a maximum of 500 frames. This preprocessing results in a total of 48K samples, with 40K samples allocated for training and 8K samples for testing.

\subsection{Baseline}

As our baseline, we train the popular temporal method, StreamMapNet, after applying the same modifications used in BEVFormer to enable training with tile images. This modified version of the model is referred to as StreamTileNet.

\subsection{Quantitative Results}

We present quantitative results for our method and the baseline. Table~\ref{results} shows the Average Precision (AP) across thresholds (0.5, 1, and 1.5 meters) for different input types (intensity and color images) and output categories (divider and boundary lines). TileTracker consistently outperforms StreamTileNet, demonstrating superior tracking. Intensity images yield better accuracy for divider lines, while differences for boundary lines are minimal, suggesting intensity images are more effective due to higher contrast. Despite using a larger dataset than nuScenes~\cite{caesar2020nuscenes} (40K compared to 28K), our results indicate that additional data and training could further improve performance due to complicated nature of our data.

\begin{table} [t]
\vspace{0mm}
  \centering
   \caption{Our model's (TileTracker) performance compared with our baseline (StreamTileNet) over different inputs (intensity and color images) and output categories (divider and boundary lines). The AP is measured for different thresholds and their average.}
   
\begin{tabular}
{p{0.8cm}p{0.8cm}p{1.5cm}p{0.7cm}p{0.7cm}p{0.7cm}p{0.5cm} }

 \hline
{\textbf{Input}}  & \textbf{Category} & \textbf{Model} &  & \textbf{AP} & & \textbf{Ave}\\
 \cline{4-6} 
 &    & & \textbf{@0.5m} & \textbf{@1.0m} & \textbf{@1.5m} & \textbf{AP}\\
\hline
 Intensity & Divider & TileTracker &  11.64 & 19.60 & 26.12 & 19.12\\
  &  & StreamTileNet &  11.95 & 18.37& 23.73 & 18.01\\
\cline{2-7}
  & Boundary & TileTracker &  2.84 & 8.74 & 15.68 & 9.09\\
  &  & StreamTileNet &  1.65 & 6.62 & 13.09 & 7.12\\
\cline{1-7}
 Color & Divider & TileTracker &   8.61 & 15.71 & 21.76 & 15.36\\
  &  & StreamTileNet &  7.57 & 14.03 & 20.29 & 13.96\\
\cline{2-7}
  & Boundary & TileTracker &   2.84 & 8.74 & 15.68 & 9.09\\
  &  & StreamTileNet &   1.79 & 6.00 & 11.90 & 6.56\\
\hline
\end{tabular}
\label{results}\vspace{-4mm}
\end{table}

\subsection{Qualitative Results}

\begin{figure*}
\begin{subfigure}{.5\textwidth}
    \centering
    \includegraphics[width=0.95\linewidth,trim={0cm  0cm 0cm 0cm },clip]{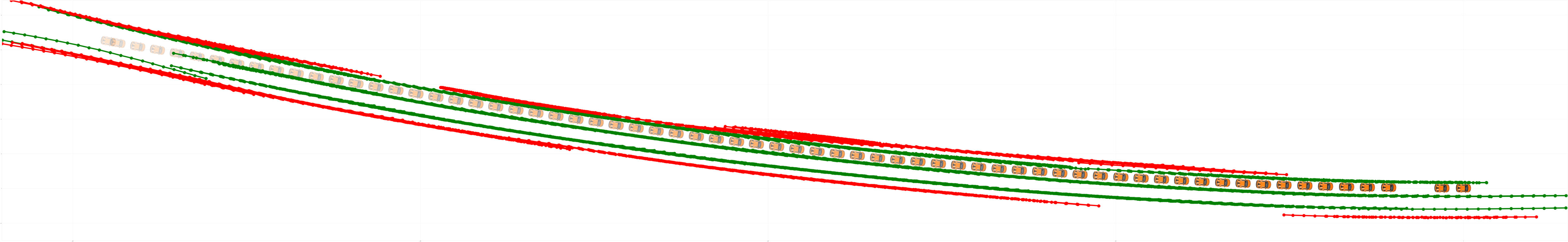}  
\end{subfigure}\hspace{0mm}
\begin{subfigure}{.5\textwidth}
    \centering
    \includegraphics[width=0.95\linewidth,trim={0cm  0cm 0cm 0cm },clip]{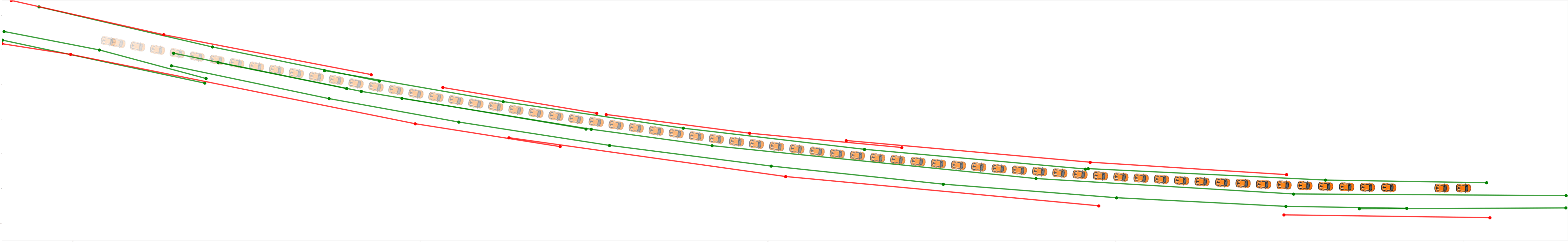}  
\end{subfigure}\hspace{0mm}
\begin{subfigure}{.5\textwidth}
    \centering
    \includegraphics[width=0.95\linewidth,trim={0cm  0cm 0cm 0cm },clip]{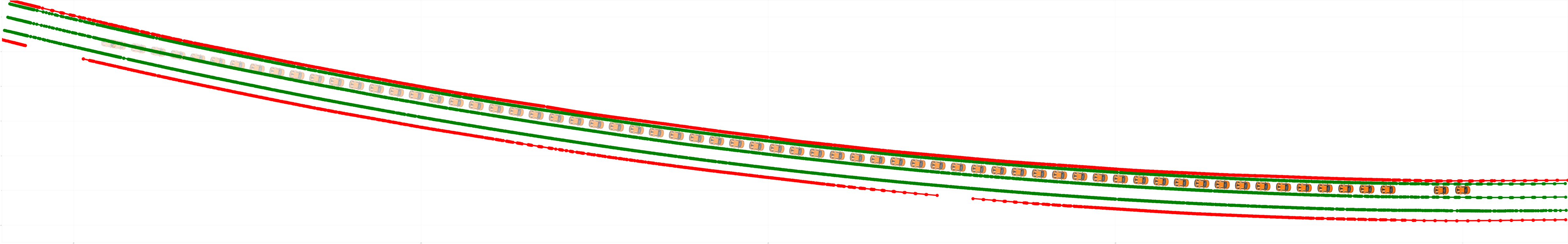}  
\end{subfigure}\vspace{0mm}
\begin{subfigure}{.5\textwidth}
    \centering
    \includegraphics[width=0.95\linewidth,trim={0cm  0cm 0cm 0.0cm },clip]{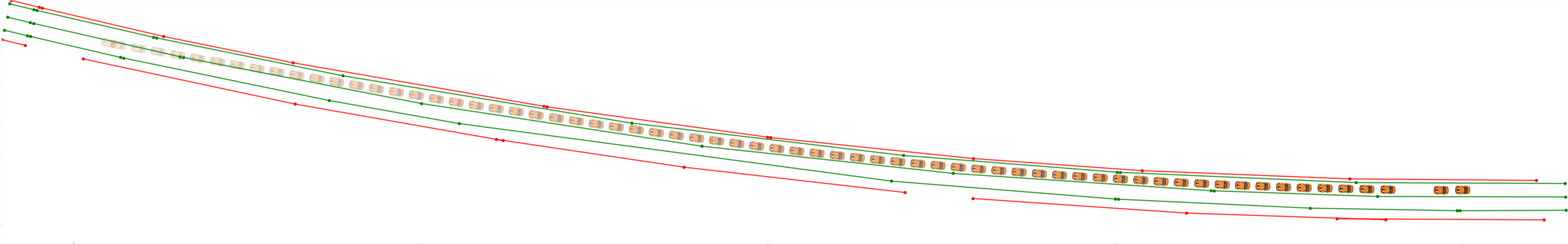}  
\end{subfigure}\hspace{0mm} 

\caption{Unmerged and merged predictions (top left and right) as well as Unmerged and merged GTs (bottom left and right) for a curved path}
\label{fig:img1}\vspace{-5mm}
\end{figure*}

\begin{figure*}
\begin{subfigure}{.5\textwidth}
    \centering
    \includegraphics[width=0.95\linewidth,trim={0cm  0cm 0cm 0cm },clip]{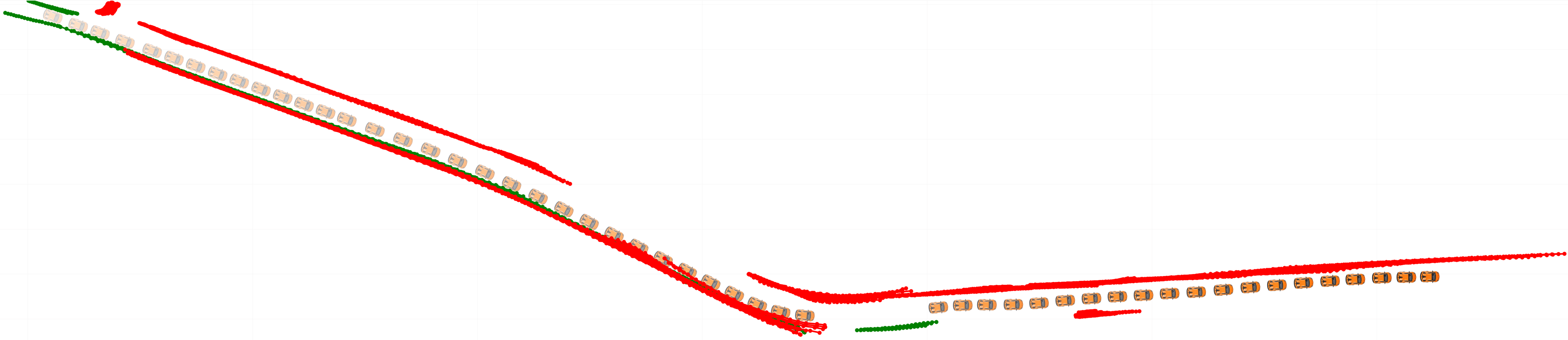}  
\end{subfigure}\hspace{0mm}
\begin{subfigure}{.5\textwidth}
    \centering
    \includegraphics[width=0.95\linewidth,trim={0cm  0cm 0cm 0cm },clip]{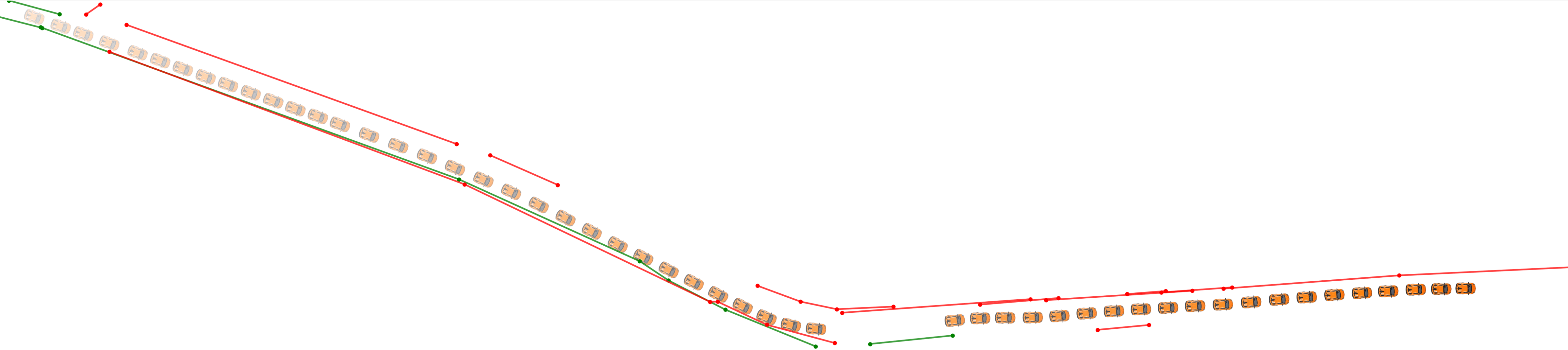}  
\end{subfigure}\hspace{0mm}
\begin{subfigure}{.5\textwidth}
    \centering
    \includegraphics[width=0.95\linewidth,trim={0cm  0cm 0cm 0cm },clip]{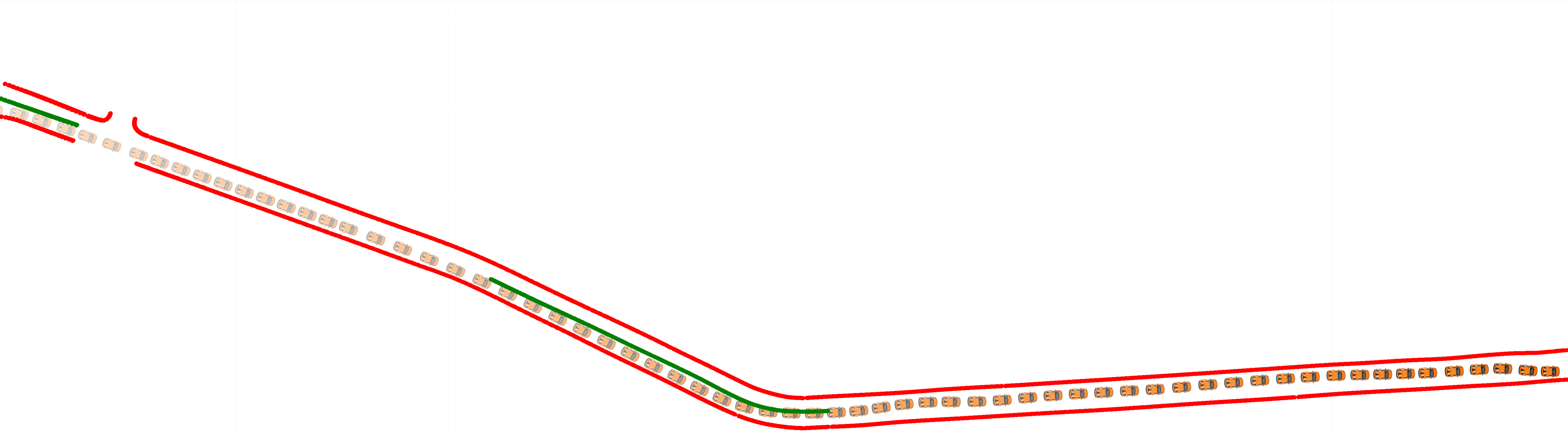}  
\end{subfigure}\vspace{0mm}
\begin{subfigure}{.5\textwidth}
    \centering
    \includegraphics[width=0.95\linewidth,trim={0cm  0cm 0cm 0.0cm },clip]{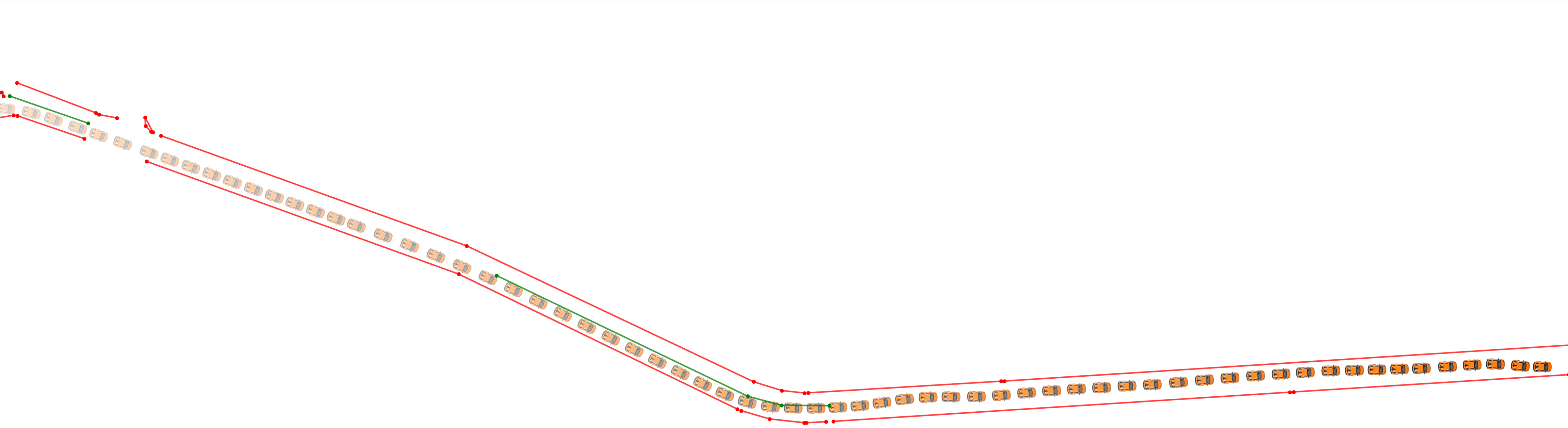}  
\end{subfigure}\hspace{0mm} 

\caption{Unmerged and merged predictions (top left and right) as well as Unmerged and merged GTs (bottom left and right) for a path containing straight paths and turns}
\label{fig:img2}\vspace{-5mm}
\end{figure*}

\begin{figure*}
\begin{subfigure}{.5\textwidth}
    \centering
    \includegraphics[width=0.95\linewidth,trim={0cm  0cm 0cm 0cm },clip]{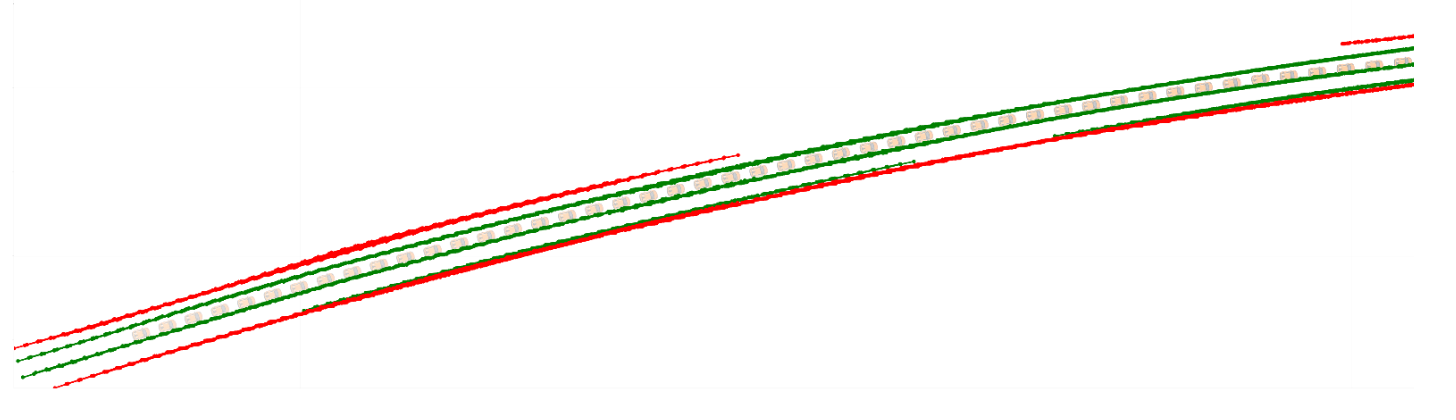}  
\end{subfigure}\hspace{0mm}
\begin{subfigure}{.5\textwidth}
    \centering
    \includegraphics[width=0.95\linewidth,trim={0cm  0cm 0cm 0cm },clip]{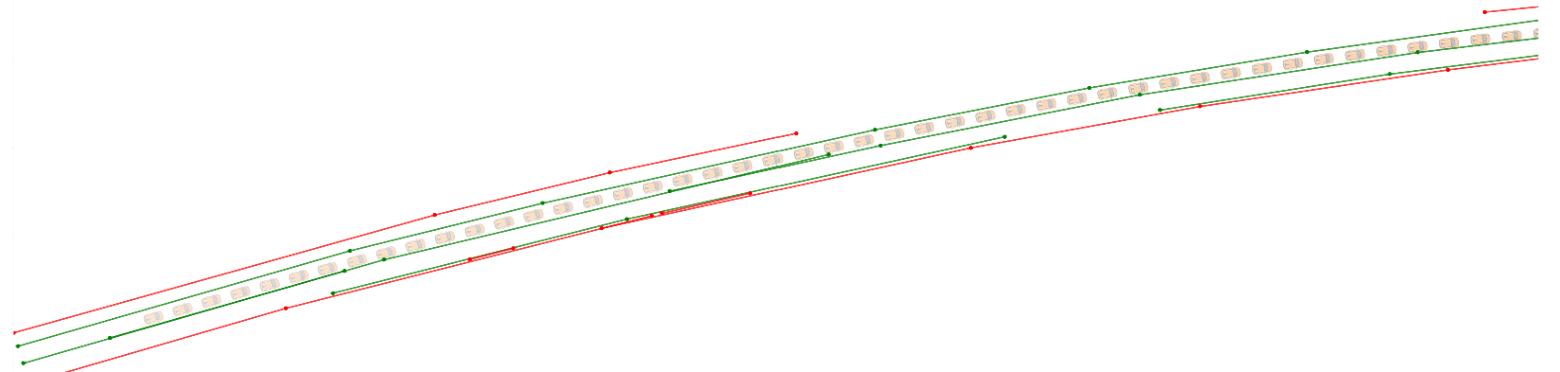}  
\end{subfigure}\hspace{0mm}
\begin{subfigure}{.5\textwidth}
    \centering
    \includegraphics[width=0.95\linewidth,trim={0cm  0cm 0cm 0cm },clip]{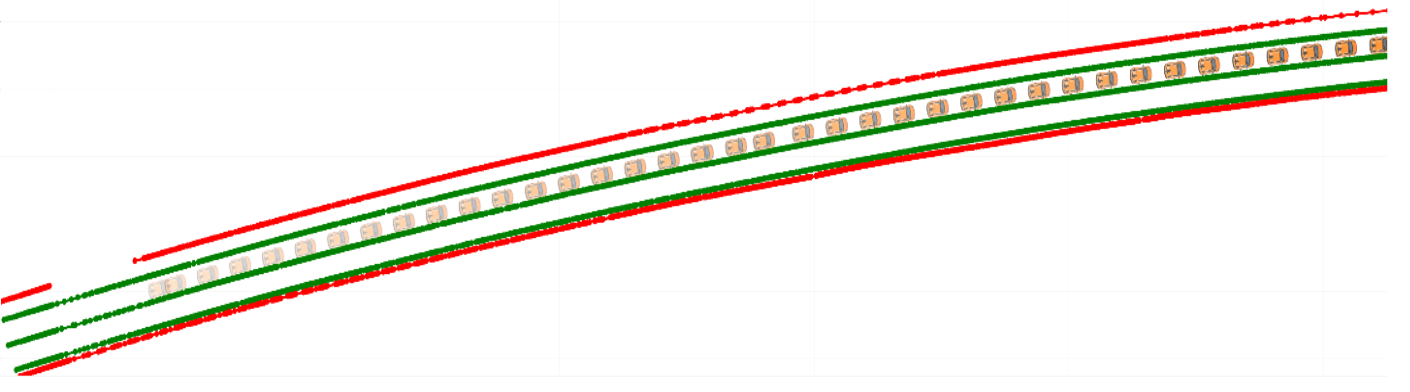}  
\end{subfigure}\vspace{0mm}
\begin{subfigure}{.5\textwidth}
    \centering
    \includegraphics[width=0.95\linewidth,trim={0cm  0cm 0cm 0.0cm },clip]{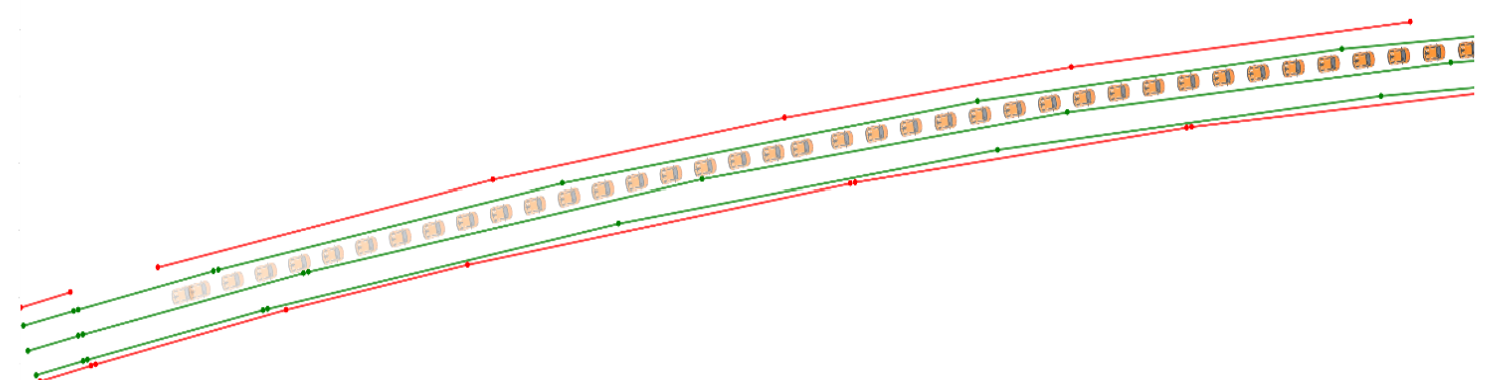}  
\end{subfigure}\hspace{0mm} 

\caption{Unmerged and merged predictions (top left and right) as well as Unmerged and merged GTs (bottom left and right) for a curved path}
\label{fig:img3}\vspace{-5mm}
\end{figure*}

\begin{figure*}
\begin{subfigure}{1\textwidth}
    \centering
    \includegraphics[width=0.95\linewidth,trim={0cm  0cm 0cm 0cm },clip]{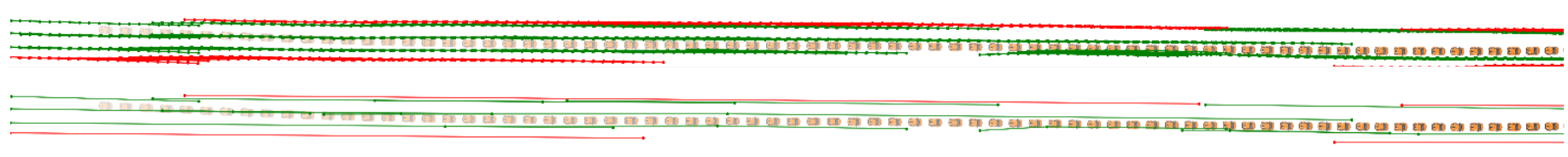}  
\end{subfigure}\hspace{0mm}
\begin{subfigure}{1\textwidth}
    \centering
    \includegraphics[width=0.95\linewidth,trim={0cm  0cm 0cm 0cm },clip]{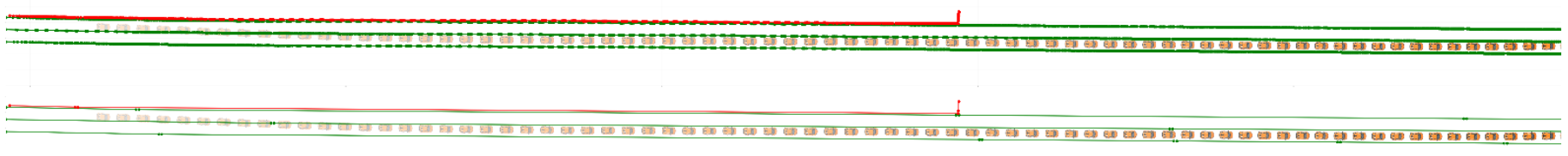}  
\end{subfigure}\hspace{0mm}
\caption{Unmerged and merged predictions (two top figures) as well as Unmerged and merged GTs (two bottom figures) for a straight path}
\label{fig:img4}\vspace{-5mm}
\end{figure*}

Qualitative results are presented in two formats: unmerged and merged. In the unmerged format, the model outputs are depicted as 20 individual vector points, plotted sequentially according to the vehicle's pose. In the merged format, the points are connected based on overlapping vectors, resulting in continuous lines. As shown in Fig.~\ref{fig:img1} to Fig.~\ref{fig:img4}, the model generally performs well. However, its performance seems better on straights and curved paths compared to sharp turns. This discrepancy is likely due to the predominance of straight and slight curved paths in our dataset. By incorporating more complex turns and curves, we can further improve the model's performance across a wider range of path types.

\vspace{-2mm}

\section{Conclusion}

In this paper, we developed a tracking-based HD mapping algorithm for top-down tile images. Building on MapTracker, we modified the BEVFormer layer to generate BEV masks from the tile images. The model was tested using both intensity and color images, with intensity images yielding better results due to the higher contrast between the lines and the background. While the model demonstrated reasonable performance, incorporating more complex paths into the dataset could further improve its effectiveness, which we leave as future work.

\vspace{-2mm}

\bibliographystyle{IEEEtran}
\bibliography{references}

\end{document}